\documentclass[10pt,twocolumn,letterpaper]{article}

\usepackage{iccv}
\usepackage{times}
\usepackage{epsfig}
\usepackage{graphicx}
\usepackage{amsmath}
\usepackage{amssymb}

\usepackage{multirow}
\usepackage{verbatim}
\usepackage{booktabs}
\usepackage{textcomp}
\usepackage{enumitem}
\usepackage{adjustbox}
\usepackage{caption}
\usepackage{subcaption}

\usepackage[breaklinks=true,bookmarks=false]{hyperref}

\iccvfinalcopy 

\def\iccvPaperID{822} 

\ificcvfinal\fi

\begin{document}

\title{VPGNet: Vanishing Point Guided Network for Lane and Road Marking Detection and Recognition}

\author{
\hspace{0mm}Seokju Lee$^\dagger$
\hspace{4.8mm}Junsik Kim$^\dagger$
\hspace{4.8mm}Jae Shin Yoon$^\dagger$
\hspace{4.8mm}Seunghak Shin$^\dagger$
\hspace{4.8mm}Oleksandr Bailo$^\dagger$
\vspace{1mm} 
\\
\hspace{0mm}Namil Kim$^\dagger$
\hspace{3.2mm}Tae-Hee Lee$^\ddagger$
\hspace{3.2mm}Hyun Seok Hong$^\ddagger$
\hspace{3.2mm}Seung-Hoon Han$^\ddagger$
\hspace{3.2mm}In So Kweon$^\dagger$
\vspace{3mm} 
\\ 
$^\dagger$Robotics and Computer Vision Lab., KAIST
\hspace{4mm}
$^\ddagger$Samsung Electronics DMC R\&D Center
\\
{\tt\small \{sjlee,jskim2,jsyoon,shshin,obailo,nikim\}@rcv.kaist.ac.kr}
\\
{\tt\small \{th810.lee,hyunseok76.hong,luoes.han\}@samsung.com, iskweon@kaist.ac.kr}
}

\maketitle
\thispagestyle{empty}

\begin{abstract}
	In this paper, we propose a unified end-to-end trainable multi-task network that jointly handles lane and road marking detection and recognition that is guided by a vanishing point under adverse weather conditions. We tackle rainy and low illumination conditions, which have not been extensively studied until now due to clear challenges. For example, images taken under rainy days are subject to low illumination, while wet roads cause light reflection and distort the appearance of lane and road markings. At night, color distortion occurs under limited illumination. As a result, no benchmark dataset exists and only a few developed algorithms work under poor weather conditions. To address this shortcoming, we build up a lane and road marking benchmark which consists of about 20,000 images with 17 lane and road marking classes under four different scenarios: no rain, rain, heavy rain, and night. We train and evaluate several versions of the proposed multi-task network and validate the importance of each task. The resulting approach, VPGNet, can detect and classify lanes and road markings, and predict a vanishing point with a single forward pass. Experimental results show that our approach achieves high accuracy and robustness under various conditions in real-time (20 fps). The benchmark and the VPGNet model will be publicly available~\footnote{
		https://github.com/SeokjuLee/VPGNet
	}.
	\vspace{-4mm}
\end{abstract}

\begin{figure}[t]
	\hspace{-3mm}
	\centering
	\captionsetup[subfigure]{aboveskip=1pt}
	\captionsetup[subfigure]{belowskip=2pt}
	\begin{tabular}{c@{\hspace{0mm}}c@{\hspace{0mm}}c@{\hspace{0mm}}c@{\hspace{0mm}}}
		\subcaptionbox{\label{intro_a}}{\includegraphics[width=0.445\linewidth]{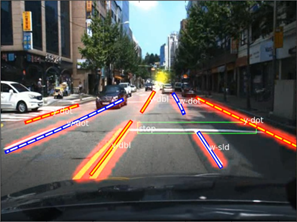}}
		{\hspace{0.1mm}}
		\subcaptionbox{\label{intro_b}}{\includegraphics[width=0.445\linewidth]{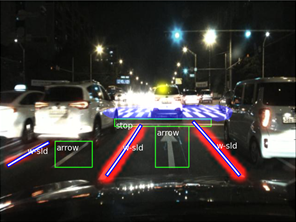}}
		\\
		\subcaptionbox{\label{intro_c}}{\includegraphics[width=0.445\linewidth]{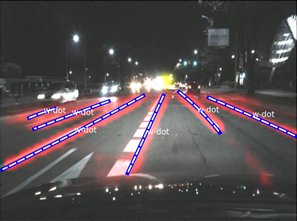}}
		{\hspace{0.1mm}}
		\subcaptionbox{\label{intro_d}}{\includegraphics[width=0.445\linewidth]{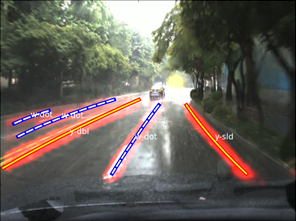}}        
	\end{tabular}
	\vspace{-4mm}
	\caption{
		Examples of our lane and road markings detection results in: (a) complex city scene; (b) multiple road markings; (c) night scene; (d) rainy condition.
		Yellow region is the vanishing area. Each class label is annotated in white.
	}
	\label{fig:intro}
	\vspace{-5mm}
\end{figure}

\vspace{-0mm}
\section{Introduction}
\label{sec:1.}
\vspace{-1mm}
Autonomous driving is a large system that consists of various sensors and control modules. The first key step for robust autonomous driving is to recognize and understand the environment around a subject. However, simple recognition of obstacles and understanding of geometry around a vehicle is insufficient. There are traffic regulations dictated by traffic symbols such as lane and road markings that should be complied with. Moreover, for an algorithm to be applicable to autonomous driving, it should be robust under diverse environments and perform in real-time.

However, research on lane and road marking detection thus far has been limited to fine weather conditions. Hand-crafted feature based methods exploit edge, color or texture information for detection, which results in a performance drop when the algorithm is tested under challenging weather and illumination conditions. Likewise, methods based on a combination of a Convolutional Neural Network (CNN) and hand-crafted features face the same challenge. Recently, a few CNN based approaches have been developed to tackle the problem in an end-to-end fashion including learning-based algorithms. They demonstrate good performance on benchmarks and in real road scenes, but are still limited to fine weather and simple road conditions.

The lack of public lane and road marking datasets is another challenge for the advancement of autonomous driving. Available datasets are often limited and insufficient for deep learning methods. For example, Caltech Lanes Dataset~\cite{aly2008real} contains 1,225 images taken from four different places. Further, Road Marking Dataset~\cite{wu2012practical} contains 1,443 images manually labeled into 11 classes of road markings. Existing datasets are all taken under sunny days with a clear scene and adverse weather scenarios are not considered.

With recent advances in deep learning, the key to robust recognition in challenging scenes is a large dataset that incorporates data captured under various circumstances.
Since no proper datasets available for lane and road marking recognition, we have collected and annotated lanes and road markings of challenging scenes captured in urban areas. Additionally, a higher network capability with a proper training scheme is required to generate a fine representation to cope with varied data. We propose to train a network that recognizes a global context in a manner similar to humans.

Interestingly, humans can drive along a lane even when it is hard to spot. 
Research works \cite{land1994we, land1995parts, salvucci2004two} have empirically shown that the driver’s gaze direction is highly correlated with the road direction. This implies that a geometric context plays a significant role in the lane localization. Inspired by this, we aim to utilize a vanishing point prediction task to embed a geometric context recognition capability to the proposed network. Further, we hope to advance autonomous driving research with the following contributions:

\begin{itemize}[noitemsep, topsep=0pt] 
	\item {\hspace{0mm}}We build up a lane and road marking detection and recognition benchmark dataset taken under various weather and illumination conditions. The dataset consists of about 20,000 images with 17 manually annotated lane and road markings classes. Vanishing point annotation is provided as well.
	\vspace{1mm}
	\item {\hspace{0mm}}We design a unified end-to-end trainable multi-task network that jointly handles lane and road marking detection and recognition that is guided by the vanishing point. We provide an extensive evaluation of our network on the created benchmark. The results show robustness under different weather conditions with real-time performance. Moreover, we suggest that the proposed vanishing point prediction task enables the network to detect lanes that are not explicitly seen.  
\end{itemize}

This paper is organized as follows. Section~\ref{sec:2.} covers recent algorithms developed for lane and road marking detection. A description of the benchmark is given in Section~\ref{sec:3.}. 
Section~\ref{sec:4.} explains our network architecture and training scheme. Experimental results are reported in Section~\ref{sec:5.}. Finally, Section~\ref{sec:6.} concludes our work.

\section{Related Work}
\label{sec:2.}
\vspace{-1mm}
In this section, we introduce previous works that aim to resolve the road scene detection challenge. Our setup as well as related works is based on a monocular vision setup.

\subsection{Lane and Road Marking Detection}
\label{sec:2.1.}
\vspace{-1mm}
Although lane and road marking detection appears to be a simple problem, the algorithm must be accurate in a variety of environments and have fast computation time.
Lane detection methods based on hand-crafted features \cite{deusch2012random,jung2013efficient,hur2013multi,borkar2012novel,satzoda2013vision,tan2014novel,wu2014lane} detect generic shapes of markings and try to fit a line or a spline to localize lanes. This group of algorithms performs well for certain situations while showing poor performance in unfamiliar conditions. In the case of road marking detection algorithms, most of the works are based on hand-crafted features. Tao \etal \cite{wu2012practical} extract multiple regions of interest as Maximally Stable Extremal Regions (MSER) \cite{matas2004robust}, and rely on FAST \cite{viswanathan2009features} and HOG\cite{dalal2005histograms} features to build templates for each road marking. Similarly, Greenhalgh \etal \cite{greenhalgh2015automatic} utilizes HOG features and a SVM is trained to produce class labels. However, as in the lane detection case, these approaches show a performance drop in unfamiliar conditions.

Recently, deep learning methods have shown great success in computer vision, including lane detection. \cite{kim2014robust, huval2015empirical} proposes a lane detection algorithm based on a CNN. Jun Li \etal \cite{li2016deep} uses both a CNN and a Recurrent Neural Network (RNN) to detect lane boundaries. In this work, the CNN provides geometric information of lane structures, and this information is utilized by the RNN that detects the lane. Bei He \etal \cite{he2016accurate} proposes using a Dual-View Convolutional Neutral Network (DVCNN) framework for lane detection. In this approach, the front-view and top-view images are fed as input to the DVCNN. Similar to the lane detection algorithms, several works have examined the application of neural networks as a feature extractor and a classifier to enhance the performance of road marking detection and recognition. 
Bailo \etal \cite{bailo2017robust} proposes a method that extracts multiple regions of interest as MSERs \cite{matas2004robust}, merges regions that possibly belong to the same class, and finally classifies region proposals by utilizing a PCANet \cite{chan2015pcanet} and a neural network.

Although the aforementioned approaches provide a promising performance of lane and road marking detection using deep learning, the problem of detection under poor conditions is still not solved. In this paper, we propose a network that performs well in any situation including bad weather and low illumination conditions.

\subsection{Object Detection by CNNs}
\label{sec:2.2.}
\vspace{-1mm}
With advances of deep learning, recognition tasks such as detection, classification, and segmentation have been solved under a wide set of conditions, yet there is no leading solution. RCNN and its variants \cite{girshick2014rich,girshick2015fast,ren2015faster} provide a breakthrough in detection and classification, outperforming previous approaches. Faster RCNN \cite{ren2015faster} replaces hand-crafted proposal methods with a convolutional network in a way that the region proposal layer shares extracted features with the classification layer. Overfeat \cite{sermanet2013overfeat} shows that a convolutional network with a sliding window approach can be efficiently computed. Its performance in object recognition and localization using multi-scale images is also reported. Some of its variants \cite{liu2015ssd,redmon2015you} achieve state of the art performance in detection tasks. Although these approaches show cutting edge results on large-scale benchmarks \cite{deng2009imagenet,everingham2010pascal,lin2014microsoft}, which contain objects that occupy a significant part of an image, the performance decreases for smaller and thinner objects~(\textit{e.g.}~lane or road markings).

Several deep learning approaches specialize in a lane and small object recognitions. For example, Huval \etal \cite{huval2015empirical} introduce a method for lanes and vehicles detection based on a fully convolutional architecture. They use the structure of \cite{sermanet2013overfeat} and extend the method with an integrated regression module composed of seven convolutional layers for feature sharing. The network is divided into two branches which perform binary classification and regression task. They evaluate results under a nice weather on a highway without complex road symbols, but do not perform a multi-label classification. 
Additionally, Zhu \etal \cite{zhu2016traffic} propose a multi-task network for traffic sign (relatively small size) detection and classification. 
In this work, the classification layer is added in parallel to the \cite{huval2015empirical} network to perform detection and classification. As a result, this work reports better performance of detecting small objects than Fast RCNN \cite{girshick2015fast}.  

\section{Benchmark}
\label{sec:3.}
\vspace{-1mm}

\subsection{Data Collection and Annotation}
\label{sec:3.1.}
\vspace{-1mm}

We have collected the dataset in various circumstances and categorized the images according to the time of the day and weather conditions. The dataset comprises situations during day time with different levels of precipitation: no rain, rainfall, and heavy rainfall.  Night time images are not subdivided by weather condition but include general images taken in a challenging situation with low illumination. The number of images for each scenario is shown in Table~\ref{weather_table}. Since our dataset is captured under bad weather conditions, we mount a camera inside a vehicle (in the center). In this way, we can avoid damaging the camera sensor while also preventing direct water drops on the camera lens. However, since several videos are recorded in heavy rain, a part of a window wiper is captured occasionally. The camera is directed to the front view of the car. Image resolution is 1288\texttimes728. Our data are captured in a downtown area of Seoul, South Korea. The shapes and symbols of the lane and road markings follow the regulations of South Korea.

We manually annotate corner points of lane and road markings. Corner points are connected to form a polygon which results in a pixel-level mask annotation for each object. In a similar manner, each pixel contains a class label. 

\begin{figure}[t]
	\begin{center}
		{\includegraphics[width=0.84\linewidth]{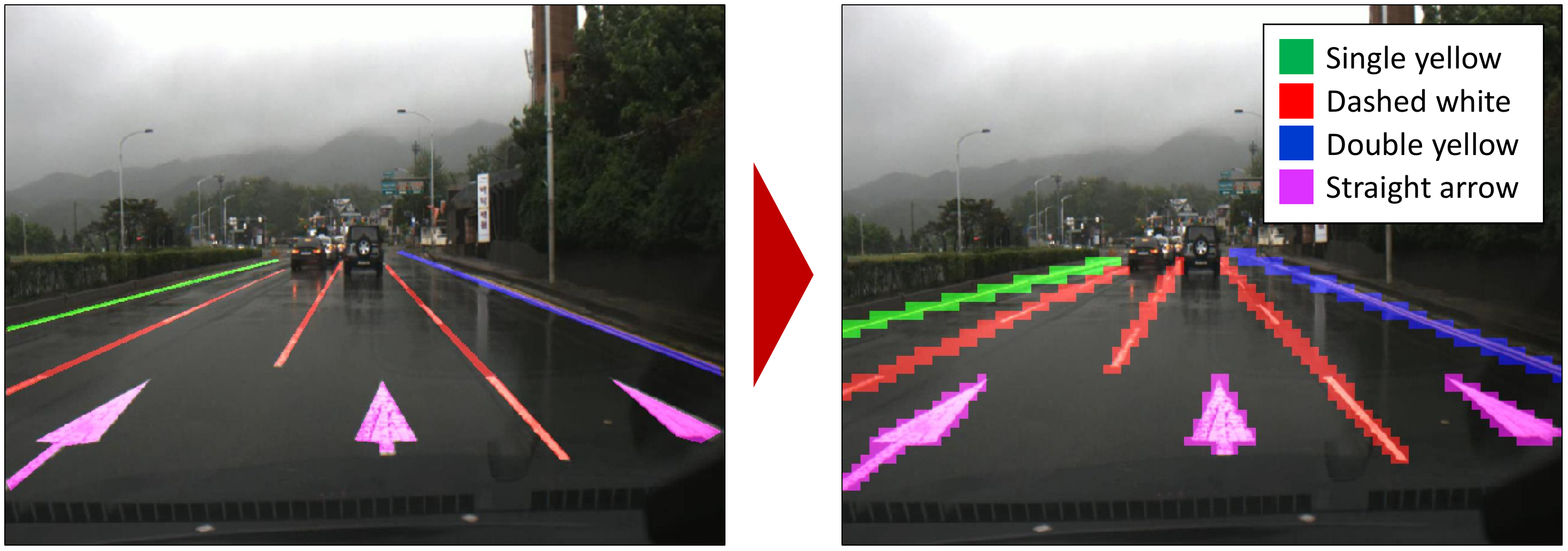}}\hspace{2mm}
	\end{center}
	\vspace{-6mm}
	\caption{Pixel- and grid-level annotations of the dataset.}
	\label{fig:annotation}
	\vspace{-6mm}
\end{figure}

However, if the network is trained with a thin lane annotation, the information tends to vanish through convolution and pooling layers. Further, since most of the neural networks require a resized image (usually smaller than original size), the thin annotations become barely visible. Therefore, we propose projecting pixel-level annotation to the grid-level mask. The image is divided into a grid 8\texttimes8 and the grid cell is filled with a class label if any pixel from the original annotation lies within the grid cell. Considering that the input size of our network is 640\texttimes480 and the output size is 80\texttimes60, the grid size is set to be proportional to the scale factor ($1/8$) between the input and output images. Specifically, the grid size is set to be 8\texttimes8. Figure~\ref{fig:annotation} shows an annotation example.

The vanishing point annotation is also provided. We localize the vanishing point in a road scene where parallel lanes supposedly meet. The vanishing point is manually annotated by a human. Depending on the scene, a difficulty level (EASY, HARD, NONE) is assigned to every vanishing point. EASY level includes a clear scene (\eg straight road); HARD level includes a cluttered scene (\eg traffic jam); NONE is where a vanishing point does not exist (\eg intersection). It is important to note that both straight and curved lanes are utilized to predict the vanishing point. We describe the definition of our vanishing point in detail in Section~\ref{sec:4.2.}. Furthermore, annotation examples are presented in the supplementary material.

\begin{table}[t]
	\centering
	\caption{Number of frames for each scenario in the dataset.}
	\vspace{-3mm}
	\label{weather_table}
	\begin{adjustbox}{width=0.4\textwidth}	
		\begin{tabular}{|c|c|c|c|c|}
			\hline
			\multicolumn{2}{|c|}{Scenario (Scn.)}          & Total frames & Training set & Test set \\ \hline \hline
			\multirow{3}{*}{Daytime} & No rain (Scn. 1)     & 13,925       & 9,184        & 4,741    \\ \cline{2-5} 
			& Rain (Scn. 2)       & 4,059        & 3,322        & 737      \\ \cline{2-5} 
			& Heavy rain (Scn. 3) & 825          & 462          & 363      \\ \hline
			\multicolumn{2}{|c|}{Night (Scn. 4)}           & 2,288        & 1,815        & 473      \\ \hline \hline
			\multicolumn{2}{|c|}{Total}                    & 21,097       & 14,783       & 6,314    \\ \hline
		\end{tabular}
	\end{adjustbox}
	\vspace{-3mm}
\end{table}

\subsection{Dataset Statistics}
\label{sec:3.2.}
\vspace{-1mm}
Our dataset consists of about 20,000 images taken during three weeks of driving in Seoul. The raw video (30 fps) is sampled at 1Hz intervals to generate image data. Images of the complex urban traffic scenes contain lane and road markings under various weather conditions during different time of the day. In total, 17 classes are annotated covering the most common markings found on the road. Although we recorded the video in various circumstances, a data imbalance between different types of lane and road markings is observed. For example, in the case of lane classes, dashed white and double yellow lines are more common than other lane types. Regarding road marking classes, straight arrows and crosswalks appear most frequently. We also define a {\textquotedblleft}Other markings{\textquotedblright} class containing road markings that are present only in South Korea, or have an insufficient number of instances to be trained as a separate class. Types of classes and the number of instances are listed in Table~\ref{table_class_stat}. 

\begin{table}[t]
	\centering
	\caption{Number of instances for each class in the dataset.}
	\vspace{-3mm}
	\label{table_class_stat}
	\begin{adjustbox}{width=0.40\textwidth}	
		\begin{tabular}{|c|c|c|c|c|c|}
			\hline
			\multicolumn{2}{|c|}{Lane} & \multicolumn{2}{c|}{Road marking} & \multicolumn{2}{c|}{Vanishing point} \\ \hline \hline
			Single white    & 25,354   & Stop line            & 7,298      & EASY             & 19,302            \\ \hline
			Dashed white    & 74,733   & Left arrow           & 1,186      & HARD             & 262               \\ \hline
			Double white    & 206      & Right arrow          & 537        & NONE             & 1,533             \\ \hline
			Single yellow   & 28,054   & Straight arrow       & 6,968      &                  &                   \\ \hline
			Dashed yellow   & 5,734    & U-turn arrow         & 127        &                  &                   \\ \hline
			Double yellow   & 8,998    & Speed bump           & 1,523      &                  &                   \\ \hline
			Dashed blue     & 1,306    & Crosswalk            & 13,632     &                  &                   \\ \hline
			Zigzag          & 1,417    & Safety zone          & 6,031      &                  &                   \\ \hline
			&          & Other markings               & 52,975      &                  &                   \\ \hline
		\end{tabular}
	\end{adjustbox}
	\vspace{-6mm}
\end{table}

\section{Neural Network}
\label{sec:4.}
\vspace{-1mm}

\begin{table*}[t]
	\centering
	\caption{Proposed network structure.}
	\vspace{-3mm}
	\label{table_net_param}
	\begin{adjustbox}{width=0.6\textwidth}	
		\begin{tabular}{|c|c|c|c|c|c|c|c|c|} 
			\hline
			Layer & Conv 1 & Conv 2 & Conv 3 & Conv 4 & Conv 5 & Conv 6 & Conv 7 & Conv 8 \\ \hline 
			\hline
			Kernel size, stride, pad & 11, 4, 0 & 5, 1, 2 & 3, 1, 1 & 3, 1, 1 & 3, 1, 1 & 6, 1, 3 & 1, 1, 0 & 1, 1, 0 \\ \hline 
			Pooling size, stride & 3, 2 & 3, 2 & \multicolumn{1}{l|}{} & \multicolumn{1}{l|}{} & 3, 2 &  &  &  \\ \hline
			Addition & LRN & LRN &  &  &  & Dropout & Dropout, branched & Branched \\ \hline
			Receptive field & 11 & 51 & 99 & 131 & 163 & 355 & 355 & 355 \\ \hline 
		\end{tabular}
	\end{adjustbox}
	\vspace{-4mm}
\end{table*}

\begin{figure*}[t]
	\begin{center}
		{\includegraphics[width=0.85\linewidth]{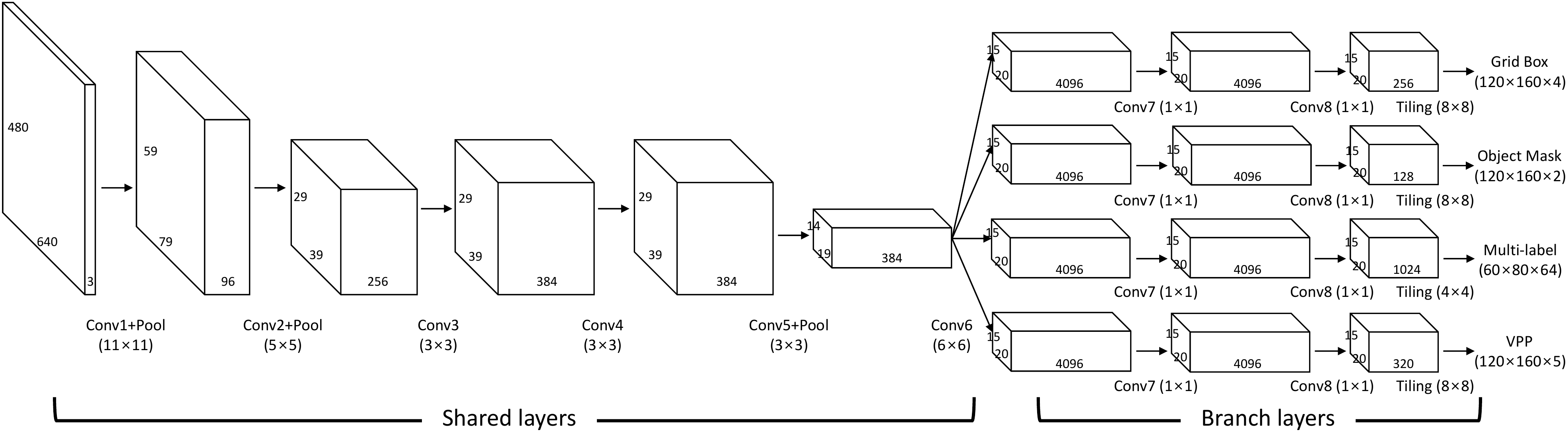}}\hspace{2mm}
	\end{center}
	\vspace{-5mm}
	\caption{VPGNet performs four tasks: grid regression, object detection, multi-label classification, and vanishing point prediction.}
	\label{fig:architecture}
	\vspace{-5mm}	
\end{figure*}

\subsection{Architecture}
\label{sec:4.1.}
\vspace{-1mm}
Our network, VPGNet, is inspired by the work of \cite{huval2015empirical} and \cite{zhu2016traffic}. The competitive advantage of our network is that it is specialized to detect and recognize lane and road markings as well as to localize vanishing point.

We propose a data layer to induce grid-level annotation that enables training of both lane and road markings simultaneously. Originally in \cite{huval2015empirical}, \cite{zhu2016traffic}, the box regression task aims to fit a single box to a particular object. This works well for objects with a blob shape (traffic signs or vehicles), but lane and road markings cannot be represented by a single box. Therefore, we propose an alternative regression that utilizes a grid-level mask. Points on the grid are regressed to the closest grid cell and combined by a multi-label classification task to represent an object. This enables us to integrate two independent targets, lane and road markings, which have different characteristics and shapes. For the post-processing, lane classes only use the output of the multi-label task, and road marking classes utilize both grid box regression and multi-label task (see Section~\ref{sec:4.4.}). Additionally, we add a vanishing point detection task to infer a global geometric context during training of patterns of lane and road markings (explained in Section~\ref{sec:4.2.}). 

The overall architecture is described in Table~\ref{table_net_param} and Figure~\ref{fig:architecture}. The network has four task modules and each task performs complementary cooperation: grid box regression, object detection, multi-label classification, and prediction of the vanishing point. This structure allows us to detect and classify the lane and road markings, and predict the vanishing region simultaneously in a single forward pass.



\subsection{Vanishing Point Prediction Task}
\label{sec:4.2.}
\vspace{-1mm}
Due to poor weather environments, illumination conditions, and occlusion, the visibility of lanes decreases. However, in such situations, humans intuitively can predict the locations of the lanes from global information such as nearby structures of roads or the flow of traffic \cite{land1994we, land1995parts, salvucci2004two}. Inspired by this, we have designed a Vanishing Point Prediction (VPP) task that guides robust lane and road marking detection similar to human vision. A vanishing point is a point where parallel lines in a three-dimensional space converge to a two-dimensional plane by a graphical perspective. In most cases of driving, lane and road markings converge to a single point regardless of whether the roads are curved or straight. In this paper, {\textquotedblleft}Vanishing Point (VP){\textquotedblright} is defined as the nearest point on the horizon where lanes converge and disappear predictively around the farthest point of the visible lane. This VP can be used to provide a global geometric context of a scene, which is important to infer the location of lanes and road markings. We integrate the VPP module with the multi-task network to train the geometric patterns of lane convergence to one point.

Borji \cite{borji2016vanishing} has shown that a CNN can localize the VP. The author vectorizes the spatial output of the network to predict the exact location of a VP by using a softmax classifier. However, selecting exactly one point over the whole network's output size results in imprecise localization. In order to provide more robust localization, we perform several experiments to guide the VP.

\begin{figure}[t]
	\hspace{-3mm}
	\centering
	\captionsetup[subfigure]{aboveskip=3pt}
	\begin{tabular}{c@{\hspace{0mm}}c@{\hspace{0mm}}}
		\subcaptionbox{\label{fig:vpp_comparison_a}}{\includegraphics[width=0.57\linewidth]{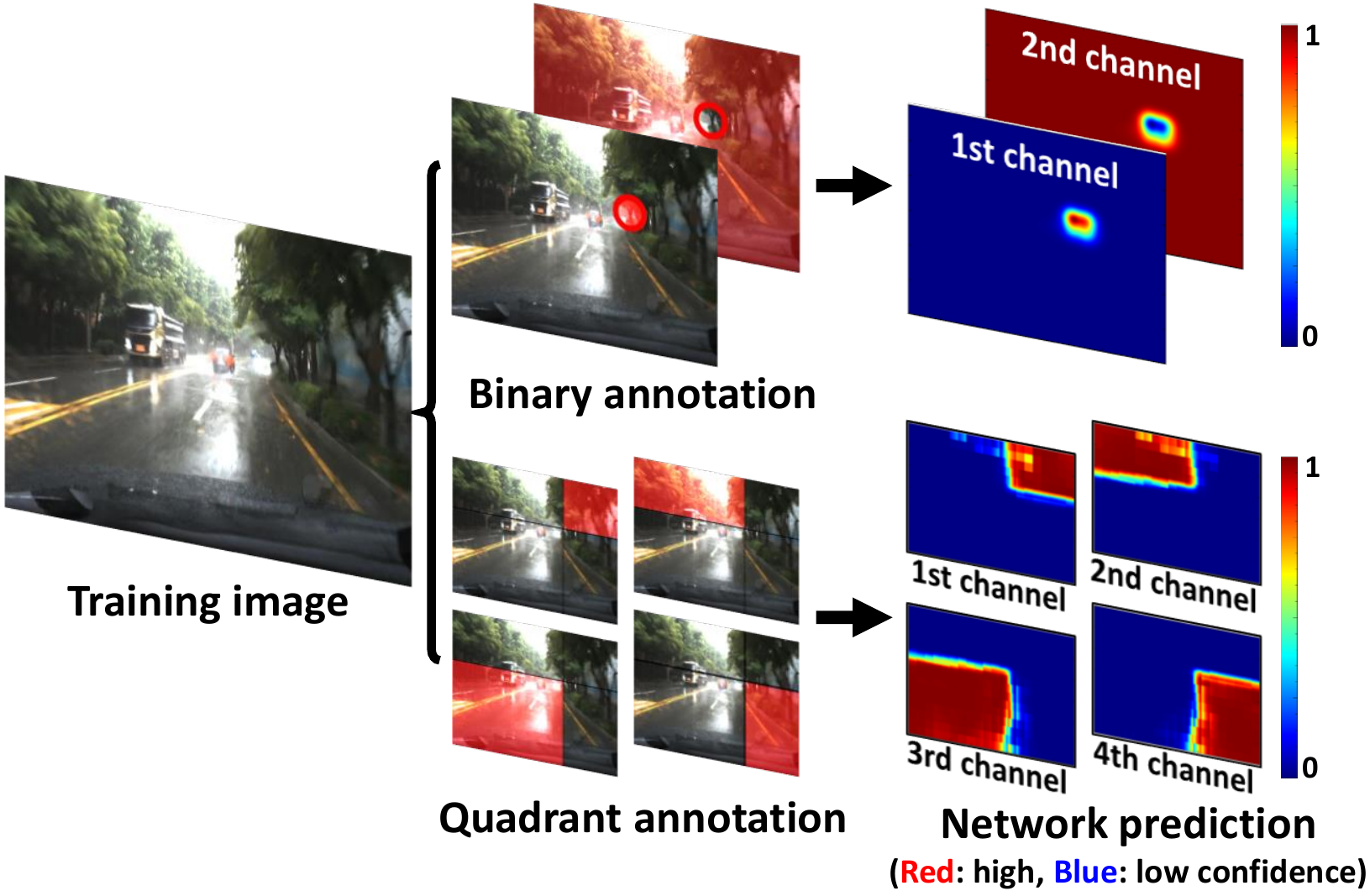}}
		\hspace{-1.8mm}
		\subcaptionbox{\label{fig:vpp_comparison_b}}{\includegraphics[width=0.42\linewidth]{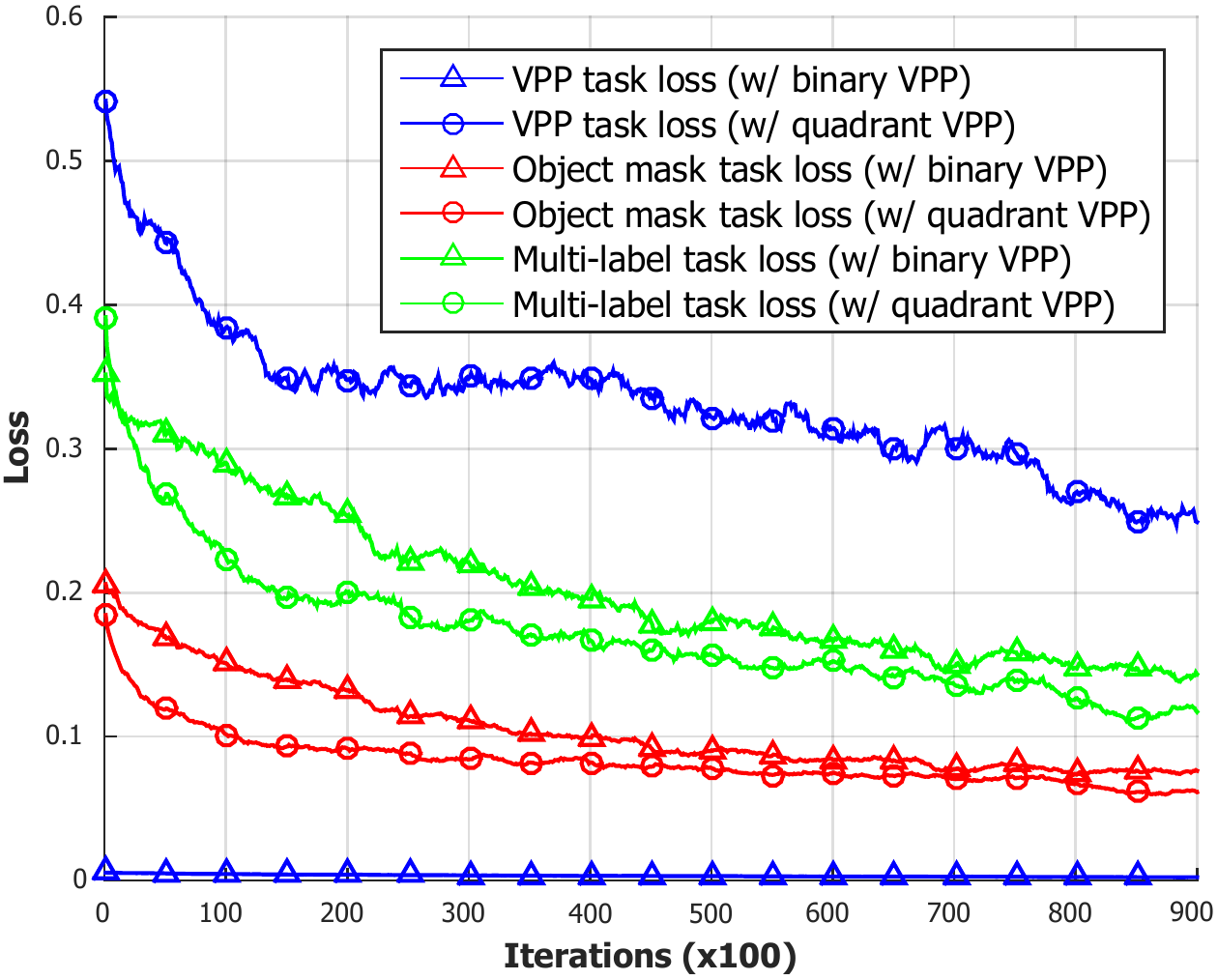}}          
	\end{tabular}
	\vspace{-3mm}
	\caption{(a) Output visualization of binary and quadrant VPP methods. For the prediction of the quadrant method, only four quadrant channels are visualized except for an absence channel. (b) The loss comparison of two methods.}
	\label{fig:vpp_comparison}
	\vspace{-5mm}
\end{figure}

First, for the VPP task, we tried regression losses~(\textit{i.e.}~L1, L2, hinge losses) that directly calculate pixel distances from a VP. Unfortunately, the results are not favorable since it is difficult to balance the losses with other tasks (object detection/multi-label classification) due to the difference in the loss scale. Therefore, we adopt a cross entropy loss to balance the gradients propagated from each of the detection tasks. By using cross entropy loss, first we apply a binary classification method that directly classifies background and foreground (~\textit{i.e.}~vanishing area, see Figure~\ref{fig:vpp_comparison_a}), as in the object detection task. The binary mask is generated in the data layer by drawing a fixed size circle centered at the VP we annotated. However, using this method on the VPP task results in extremely fast convergence of the training loss. This is caused by the imbalance of the number of background and foreground pixels. Since the vanishing area is drastically smaller than the background, the network is initialized to infer every pixel as background class. This phenomenon contradicts our original intention of training the VPP to learn the global context of a scene.

Considering the challenge imposed by the aforementioned imbalance during the binary VPP method, we have newly designed the VPP module. As stated before, the purpose of attaching the VPP task is to improve a scene representation that implies a global context to predict invisible lanes due to occlusions or extreme illumination condition. The whole scene should be taken into account to efficiently reflect global information inferring lane locations. We use a quadrant mask that divides the whole image into four sections. The intersection of these four sections is a VP. In this way, we can infer the VP using four quadrant sections which cover the structures of a global scene. To implement this, we define five channels for the output of the VPP task: one absence channel and four quadrant channels. Every pixel in the output image chooses to belong to one of the five channels. The absence channel is used to represent a pixel with no VP, while the four quadrant channels stand for one of the quadrant sections on the image. For example, if the VP is present in the image, every pixel should be assigned to one of the quadrant channels, while the absence channel cannot be chosen. Specifically, the third channel would be guided by the upper right diagonal edges from the road scene, and the fourth channel would extract the upper left diagonal edges from the road scene. On the other hand, if the VP is hard to be identified (\textit{e.g.}~intersection roads, occlusions), every pixel will tend to be classified as the absence channel. In this case, the average confidence of the absence channel would be high.

Unlike the binary classification approach, our quadrant method enriches the gradient information that contains a global structure of a scene. The loss comparison in Figure~\ref{fig:vpp_comparison_b} indirectly shows that the network is trained without overfitting compared to the binary case. Note that we only use the quadrant VPP method for the evaluation. The binary VPP method is introduced only to show readers that a naive VPP training scheme does not yield satisfactory results. The whole multi-task network allows us to detect and recognize the lane and road marking, as well as to predict the VP simultaneously in a single forward pass.


\subsection{Training}
\label{sec:4.3.}
\vspace{-1mm}

Our network includes four tasks which cover different contexts. The detection task recognizes objects and covers a local context, while the VPP task covers a global context. If those tasks are trained altogether at the same training phase, the network can be highly influenced by a certain dominant task. We noticed that during the training stage the VPP task became dependent on the lane detection task. The dependency between lanes and the VP implies a strong information correlation. In this case, the VP provides redundant information to the network, leading to marginal lane detection improvement. In order to prevent this side effect, we train the network in two phases to tolerate the balance between the tasks.

In the first phase, we train only the VPP task. We fix the learning rates to zero for every task except the VPP module. In this way, we can train the kernels to learn a global context of the image. The training of this phase stops upon reaching convergence of the VP detection task. Although we train only the VPP task, due to the weight update of the mutually shared layers, losses of the other detection tasks are also decreased by about 20\%. This shows that lane and road marking detection and VPP tasks share some common characteristics in the feature representation layers.

In the second phase, we further train all the tasks using the initialized kernels from the first phase. Since all tasks are trained together at this point, it is important to balance their learning rates. If a certain task loss weight is small, it becomes dependent on other tasks and vice versa. Equation (\ref{eq_1}) shows the summation of four losses from each task:
\begin{equation}
\vspace{-1mm}
\small
\label{eq_1}
{\scriptsize
	Loss = {w_1}{L_{reg}} + {w_2}{L_{om}} + {w_3}{L_{ml}} + {w_4}{L_{vp}}
}
\end{equation}
where ${L_{reg}}$ is a grid regression L1 loss, ${L_{om}}$ and ${L_{ml}}$ and ${L_{vp}}$ are cross entropy losses in each branch of the network. We balance the tasks by weight terms ${w_{1}} {\sim} {w_{4}}$ in the following way. First, ${w_{1}} {\sim} {w_{4}}$ are set to be equal to ${1}$, and the starting losses are observed. 
Then, we set the reciprocal of these initial loss values to the loss weight so that the losses are uniform. In the middle of the training, if the scale difference between losses becomes large, this process is repeated to balance the loss values. The second phase stops when the validation accuracy is converged.

\subsection{Post-Processing}
\label{sec:4.4.}
\vspace{-1mm}

Each lane and road marking class and VPs are required to be represented suitably for real world application. Therefore, we implement post-processing techniques to generate visually satisfying results.

\textbf{Lane} ~ In the case of the lane classes, we use the following techniques: point sampling, clustering, and lane regression. First, we subsample local peaks from the region where the probability of lane channels from the multi-label task is high. The sampled points are potential candidates to become the lane segments. Further, selected points are projected to the bird’s-eye view by inverse perspective mapping (IPM) \cite{bertozzi1996real}. IPM is used to separate the sampled points near the VP. This is useful not only for the case of straight roads but also curved ones. We then cluster the points by our modified density-based clustering method. We sequentially decide the cluster by the pixel distance. After sorting the points by the vertical index, we stack the point in a bin if there is a close point among the top of the existing bins. Otherwise, we create a new bin for a new cluster. By doing this, we can reduce the time complexity of the clustering. The last step is quadratic regressions of the lines from the obtained clusters utilizing the location of the VP. If the farthest sample point of each lane cluster is close to the VP, we include it in the cluster to estimate a polynomial model. This makes the lane results stable near the VP. The class type is assigned to each line segment from the multi-labeled output of the network.

\textbf{Road marking} ~ For the road marking classes, grid sampling and box clustering are applied. First, we extract grid cells from the grid regression task with high confidence for each class from the multi-label output. We then select corner points of each grid and merge them with the nearby grid cells iteratively. If no more neighboring grid cells belong to the same class, the merging is terminated. Some road markings such as crosswalks or safety zones that are difficult to define by a single box are localized by grid sampling without subsequent merging. 

\textbf{Vanishing point} ~ 
Our VPP module outputs five channels of the confidence map: four quadrant channels and one absence channel. Through these quadrants, we generate the location of a VP. The VP is where all four quadrants intersect. That is, we need to find a point where four confidences from each quadrant channel become close. Equation (\ref{eq_2}) and (\ref{eq_3}) describe the boundary intersection of each quadrant:
\begin{equation}
\vspace{-2mm}
\small
\label{eq_2}
{P_{avg}} = \frac{{1 - (\sum {p_{0}}(x,y))/(m \times n)}}{4}
\end{equation}
\begin{equation}
\small
\label{eq_3}
{loc}_{vp}  = \mathop {\arg\min}\limits_{(x,y)} \sum\limits_{n = 1}^4 {{|{P_{avg}} - {p_{n}}(x,y)|^2}}
\end{equation}
where ${P_{avg}}$ is the probability that a VP exists in the image, ${p_{n}}(x,y)$ is the confidence of ${(x,y)}$ on ${n_{th}}$ channel (${n=0}$:~\textit{absence channel}), $m$$\times$$n$ is the confidence map size, and ${loc}_{vp}$ is the location of the VP.

\section{Results}
\label{sec:5.}
\vspace{-1mm}
Our experiments consist of six parts. First, we show the experimental settings such as dataset splits and training parameters. Secondly, we provide an analysis of our network. We explore how multiple tasks jointly cooperate and affect the performance of each other. Third, our evaluation metric for each target is introduced. Lastly, we show lanes, road markings, and VPs detection and classification results.

\subsection{Experimental Settings}
\label{sec:5.1.}
\vspace{-1mm}
A summary of the datasets is provided in Table~\ref{weather_table}. During the training, we double the number of images by flipping the original ones. This, in turn, doubles the training set and also prevents positional bias that comes from the lane positions. More specifically, the dataset is obtained in a right-sided driving country, and by flipping the dataset we can simulate a left-sided environment. 

At the first training phase, we initialize the network only by the VPP task. After the initialization, all four tasks are trained simultaneously. For every task, we use Stochastic Gradient Descent optimization with a momentum of 0.9 and a mini-batch size of 20. Since multiple tasks must converge proportionally, we tune the learning rate of each task. 

We train three models of the network divided by task: 2-Task (revised \cite{huval2015empirical}), 3-Task (revised \cite{zhu2016traffic}), and 4-Task (VPGNet). 2-Task network includes regression and binary classification tasks. 3-Task network includes 2-Task and a multi-label classification task. 4-Task network includes 3-Task and a VPP task, which is the VPGNet. 
Since the lane detection in \cite{huval2015empirical} is not fully reproducible, we modify the data layer to handle the grid mask and move one convolutional layer from shared layers to branch layers, as in the 3- and 4-Task networks. The 3-Task network is similar to \cite{zhu2016traffic}, but we modify the data layer to handle the grid mask.

We test our models on NVIDIA GTX Titan X and achieve a speed of 20 Hz by using only a single forward pass. Specifically, the single forward pass takes about 30 ms and the post-processing takes about 20 ms or less.

\subsection{Analysis of Multi Task Learning}
\label{sec:5.2.}
\begin{figure}[t]
	\begin{center}
		{\includegraphics[width=0.95\linewidth]{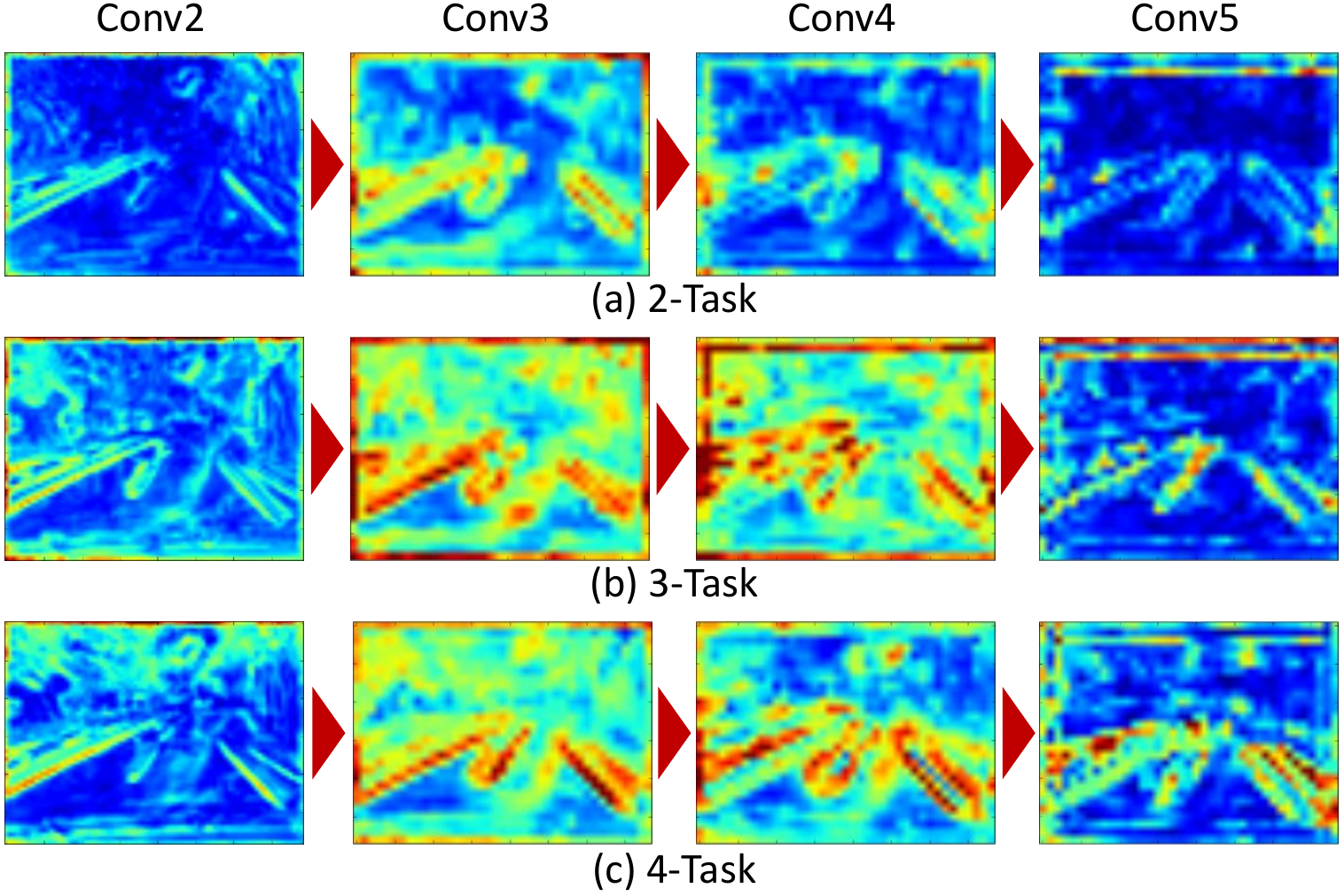}}\hspace{2mm}
	\end{center}
	\vspace{-4mm}
	\caption{Activated neurons in the feature sharing network. Intensity scale in each layer activation is equalized.}
	\label{fig:activation}
	\vspace{-4mm}
\end{figure}

\vspace{-1mm}
In this section, we validate whether our multi-task modules contribute to improvement of the network training. We observe the activated neurons in the feature sharing network. From the lower to higher layer, the abstraction level is accelerated. Figure~\ref{fig:activation} shows the activated neurons after each convolutional layer before the branch. We average over all channel values. For a fair comparison, we equalize the intensity scale in each layer activation. As the results show, if we use more tasks, more neurons respond, especially around the boundaries of roadways.

\subsection{Evaluation Metrics}
\label{sec:5.3.}
\vspace{-1mm}
In this section, we show the newly proposed evaluation metrics for our benchmark evaluation. First, we introduce our evaluation metric for the lane detection. Since the ground truth of our benchmark is annotated with grid cells we compute the minimum distance from the center of each cell to the sampled lane points for every cell. If the minimum distance is within the boundary ${R}$, we mark these sampled points as true positive and the corresponding grid cell as detected. By measuring every grid cell on the lane, we can strictly evaluate the location of lane segments. Additionally, we measure F1 score for the comparison.

In the case of road markings, we use mitigated evaluation measurement. Since the only information we need while driving is the road marking in front of us rather than the exact boundary of the road markings, we measure the precision of predicted blobs. Specifically, we count all predicted cells overlapped with the ground truth grid cells. The overlapped cells are marked as true positive cells. If the number of true positive cells is greater than half of the number of all predicted cells over a clustered blob, the overlaid ground truth target is defined as detected. Additionally, we measure the recall score for comparison.

\begin{figure}[t]
	\begin{center}
		{\includegraphics[width=0.74\linewidth]{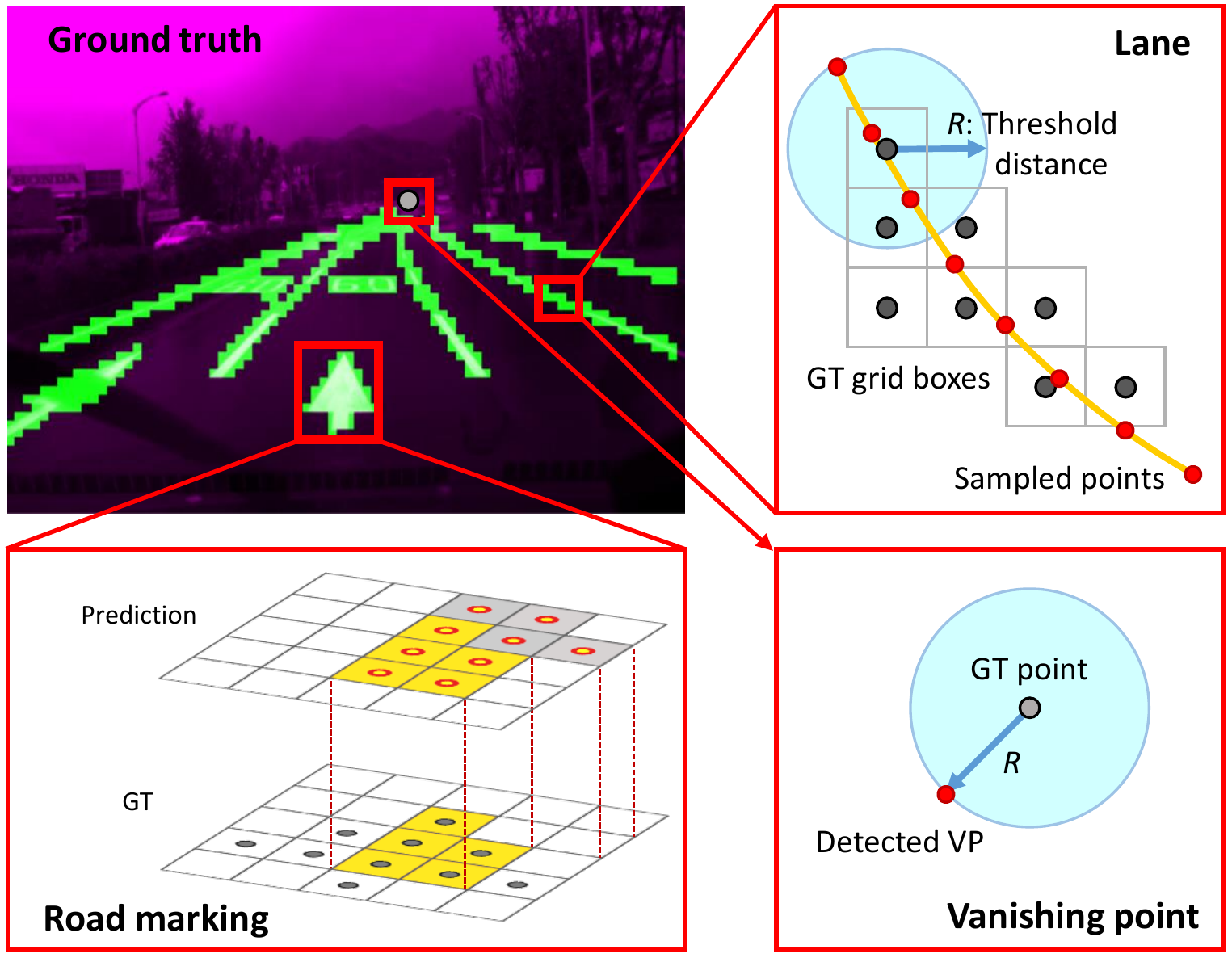}}\hspace{2mm}
	\end{center}
	\vspace{-5mm}
	\caption{Graphical explanation of the evaluation metrics.}
	\label{fig:eval}
	\vspace{-4mm}
\end{figure}

For evaluation of the VP, we measure the Euclidean distance between a ground truth point and a predicted VP. The recall score is evaluated by varying the threshold distance ${R}$ from the ground truth VP. Figure~\ref{fig:eval} shows a summary of how we measure all three targets of our network.

\subsection{Lane Detection and Recognition}
\label{sec:5.4.}
\begin{figure}[b]
	\vspace{-4mm}
	\begin{center}
		{\includegraphics[width=0.90\linewidth]{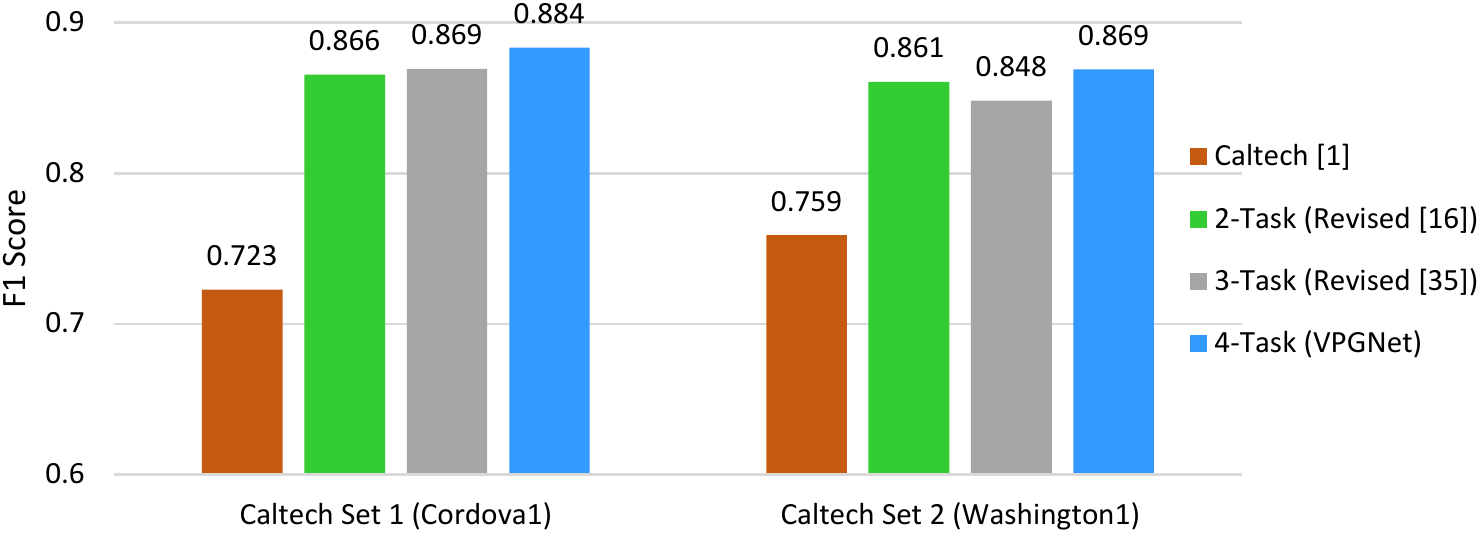}}\hspace{2mm}
	\end{center}
	\vspace{-5.5mm}
	\caption{Lane detection score on Caltech lanes dataset.}
	\label{fig:caltech}
\end{figure}

\vspace{-1mm}
For lane classes, we measure detection, as well as simultaneous detection and classification performance. First, we compare our multi-task networks with the baseline methods in the Caltech Lanes Dataset~\cite{aly2008real} (see Figure~\ref{fig:caltech}). We set ${R}$ to equal to the average half value of the lane thickness (20 pixels). Due to perspective effect, the double lane in front of the camera is about 70 to 80 pixels thick, and it is as small as 8 pixels (a single grid size) near the VP. Since this dataset contains relatively easy scenes during daytime, the overall performance of 2-, 3-, and 4-Task networks is very similar. Nevertheless, our network achieves the best F1 score.

\begin{figure}
	\begin{center}
		{\includegraphics[width=0.95\linewidth]{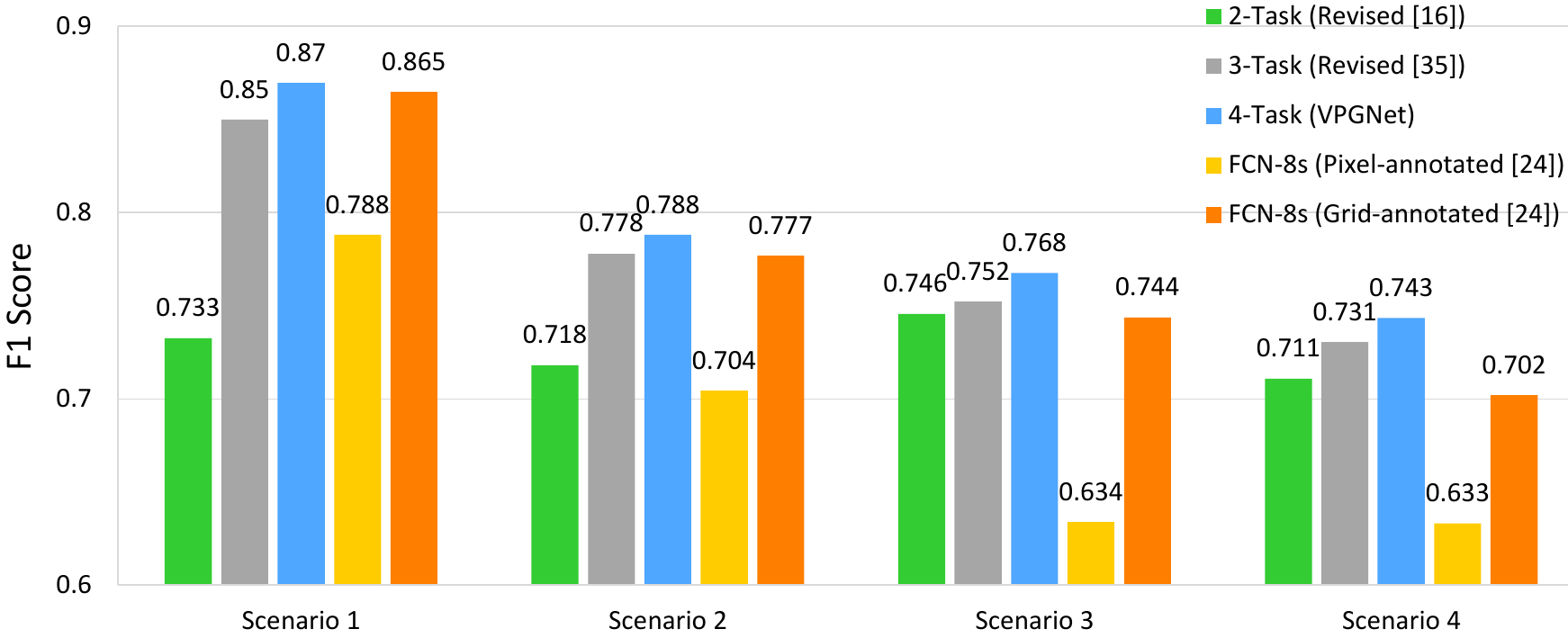}}\hspace{2mm}
	\end{center}
	\vspace{-5mm}
	\caption{Lane detection score on our benchmark.}
	\label{fig:lane_detection}
	\vspace{-5mm}
\end{figure}

Further, we provide a comparison of the proposed three versions of multi-task networks and the FCN-8s \cite{long2015fully} segmentation method on our benchmark dataset. It is important to note that our networks utilize grid-level annotation, while FCN-8s is trained independently with both pixel- and grid-level annotations. For testing purposes, four scenarios have been selected as in Section~\ref{sec:5.1.}, and the F1 score is compared in each scenario. Figure~\ref{fig:lane_detection} shows the experimental results. Noticeably, our method shows significantly better lane detection performance in each bad weather condition scenario. Moreover, the forward pass time of the VPGNet is 30 ms, while FCN-8s~\cite{long2015fully} takes 130 ms.

Interestingly, FCN-8s shows better performance with the proposed grid-level annotation scheme compared to pixel-level annotation. This proves that the grid-level annotation is more suitable for lane detection and recognition. The reason is that grid-level annotation generates stronger gradients from the edge information around the thinly annotated area (\textit{i.e.}~lane or road markings), which, in turn, results in enriched training and leads to better performance.

In order to see what happens if the VP does not exist, we conducted an additional test on images without the VP (\eg intersection roads or occlusions). Table~\ref{novp} shows the results of the experiment, demonstrating that the enhancement of feature representation through the VPP task helps to find lanes even when there is no VP. Selected results are shown in the supplementary material.

\begin{table}[b]
	\vspace{-4mm}
	\footnotesize
	\centering
	\caption{Lane detection score on No-VP set
		(\textcolor{red}{\textbf{Red}}: Best).}
	\vspace{-2.5mm}
	\label{novp}
	\begin{adjustbox}{width=0.38\textwidth}
		\begin{tabular}{|c|c|c|c|c|}
			\hline
			& \begin{tabular}[c]{@{}c@{}}FCN-8s\\ (pixel)\end{tabular} & \begin{tabular}[c]{@{}c@{}}FCN-8s\\ (grid)\end{tabular} & \begin{tabular}[c]{@{}c@{}}3-Task\\ (revised~\cite{zhu2016traffic})\end{tabular} & \begin{tabular}[c]{@{}c@{}}4-Task\\ (VPGNet)\end{tabular}  \\ \hline \hline
			No-VP set & 0.3310                                       & 0.4496                                                   & 0.4535                                                   & \textcolor{red}{0.5234}                                  \\ \hline
		\end{tabular}
	\end{adjustbox}
	\vspace{-1mm}
\end{table}

For the simultaneous detection and classification of lane classes, due to the class imbalance, we measure the F1 score of the top four lane classes by the number of instances. The selected classes are: single white, dashed white, single yellow, and double yellow lines. Table~\ref{lane_classification} shows the performance of the 3- and 4-Task networks. Except for {\textquotedblleft}no rain, daytime condition{\textquotedblright}, recognition of the single white line is highly improved. This shows that using the VPP task on rainy and night conditions improves the activation of roadway boundaries which are usually marked with single white lines.

\begin{table}[b]
	\vspace{-1mm}
	\footnotesize
	\centering
	\caption{Simultaneous detection and classification F1 score for lane classes 
		(\textcolor{red}{\textbf{Red}}: Best).}
	\vspace{-2.5mm}
	\label{lane_classification}
	\begin{adjustbox}{width=0.44\textwidth}
		\begin{tabular}{|l|c|c|c|c|c|}
			\hline
			\multicolumn{2}{|c|}{Lane class} & \begin{tabular}[c]{@{}c@{}}Single white\end{tabular} & \begin{tabular}[c]{@{}c@{}}Dashed white\end{tabular} & \begin{tabular}[c]{@{}c@{}}Single yellow\end{tabular} & \begin{tabular}[c]{@{}c@{}}Double yellow\end{tabular} \\ \hline \hline
			\multirow{2}{*}{Scenario 1} & 3-Task & \textcolor{red}{0.55}                                                  & \textcolor{red}{0.77}                                                   & 0.57                                                   & 0.32                                                    \\
			& 4-Task & 0.49                                                  & 0.76                                                   & \textcolor{red}{0.58}                                                   & \textcolor{red}{0.36}                                                    \\ \hline
			\multirow{2}{*}{Scenario 2} & 3-Task & 0.45                                                  & \textcolor{red}{0.67}                                                   & 0.64                                                   & \textcolor{red}{0.62}                                                    \\
			& 4-Task & \textcolor{red}{0.52}                                                 & 0.66                                                   & \textcolor{red}{0.65}                                                   & 0.61                                                    \\ \hline
			\multirow{2}{*}{Scenario 3} & 3-Task & 0.31                                                  & 0.72                                                   & 0.70                                                   & 0.37                                                    \\
			& 4-Task & \textcolor{red}{0.42}                                                  & \textcolor{red}{0.73}                                                   & \textcolor{red}{0.71}                                                   & \textcolor{red}{0.40}                                                    \\ \hline
			\multirow{2}{*}{Scenario 4} & 3-Task & 0.27                                                  & 0.68                                                   & \textcolor{red}{0.48}                                                   & 0.36                                                    \\
			& 4-Task & \textcolor{red}{0.42}                                                  & \textcolor{red}{0.69}                                                   & 0.42                                                   & \textcolor{red}{0.40}                                                    \\ \hline
		\end{tabular}
	\end{adjustbox}
\end{table}

\subsection{Road Marking Detection and Recognition}
\label{sec:5.5.}
\vspace{-1mm}

In the case of road marking classes, we evaluate the simultaneous detection and classification performance. Due to the dataset imbalance of road marking classes, we measure the recall score of the top four road marking classes by the number of instances. The selected classes are as follows: stop line, straight arrow, crosswalk, and safety zone. Table~\ref{rm_classification} shows the performance of 3- and 4-Task networks. Except for the stop line class in {\textquotedblleft}no rain, daytime condition{\textquotedblright}, the evaluation results are highly improved. This makes sense because the stop line has horizontal edges which are not closely related to the VPP task. Other road markings have shapes that give directions to VP from a geometric perspective. Consequently, responses to those classes become highly activated.

\begin{table}[t]
	\footnotesize
	\centering
	\caption{Simultaneous detection and classification recall score for road marking classes (\textcolor{red}{\textbf{Red}}: Best).}
	\vspace{-2mm}
	\label{rm_classification}
	\begin{adjustbox}{width=0.44\textwidth}
		\begin{tabular}{|l|c|c|c|c|c|}
			\hline
			\multicolumn{2}{|c|}{\begin{tabular}[c]{@{}c@{}}Road marking class\end{tabular}} & \begin{tabular}[c]{@{}c@{}}Stop line\end{tabular} & \begin{tabular}[c]{@{}c@{}}Straight arrow\end{tabular} & Crosswalk & \begin{tabular}[c]{@{}c@{}}Safety zone\end{tabular} \\ \hline \hline
			\multirow{2}{*}{Scenario 1}                            & 3-Task                            & \textcolor{red}{0.83}                                                & 0.46                                                     & 0.88      & 0.59                                                  \\
			& 4-Task                            & 0.78                                                & \textcolor{red}{0.80}                                                     & \textcolor{red}{0.94}      & \textcolor{red}{0.80}                                                  \\ \hline
			\multirow{2}{*}{Scenario 2}                            & 3-Task                            & 0.60                                                & 0.41                                                     & 0.81      & 0.47                                                  \\
			& 4-Task                            & \textcolor{red}{0.73}                                                & \textcolor{red}{0.65}                                                     & \textcolor{red}{0.85}      & \textcolor{red}{0.65}                                                  \\ \hline
			\multirow{2}{*}{Scenario 3}                            & 3-Task                            & 0.33                                                & 0.39                                                     & 0.84      & 0.47                                                  \\
			& 4-Task                            & \textcolor{red}{0.56}                                                & \textcolor{red}{0.63}                                                     & \textcolor{red}{0.93}      & \textcolor{red}{0.61}                                                  \\ \hline
			\multirow{2}{*}{Scenario 4}                            & 3-Task                            & 0.60                                                & 0.48                                                     & 0.82      & 0.37                                                  \\
			& 4-Task                            & \textcolor{red}{0.80}                                                & \textcolor{red}{0.68}                                                     & \textcolor{red}{0.89}      & \textcolor{red}{0.38}                                                  \\ \hline
		\end{tabular}
	\end{adjustbox}
	\vspace{-1mm}
\end{table}

\subsection{Vanishing Point Prediction}
\label{sec:5.6.}
\vspace{-1mm}

\begin{figure}[t]
	\begin{center}
		{\includegraphics[width=0.95\linewidth]{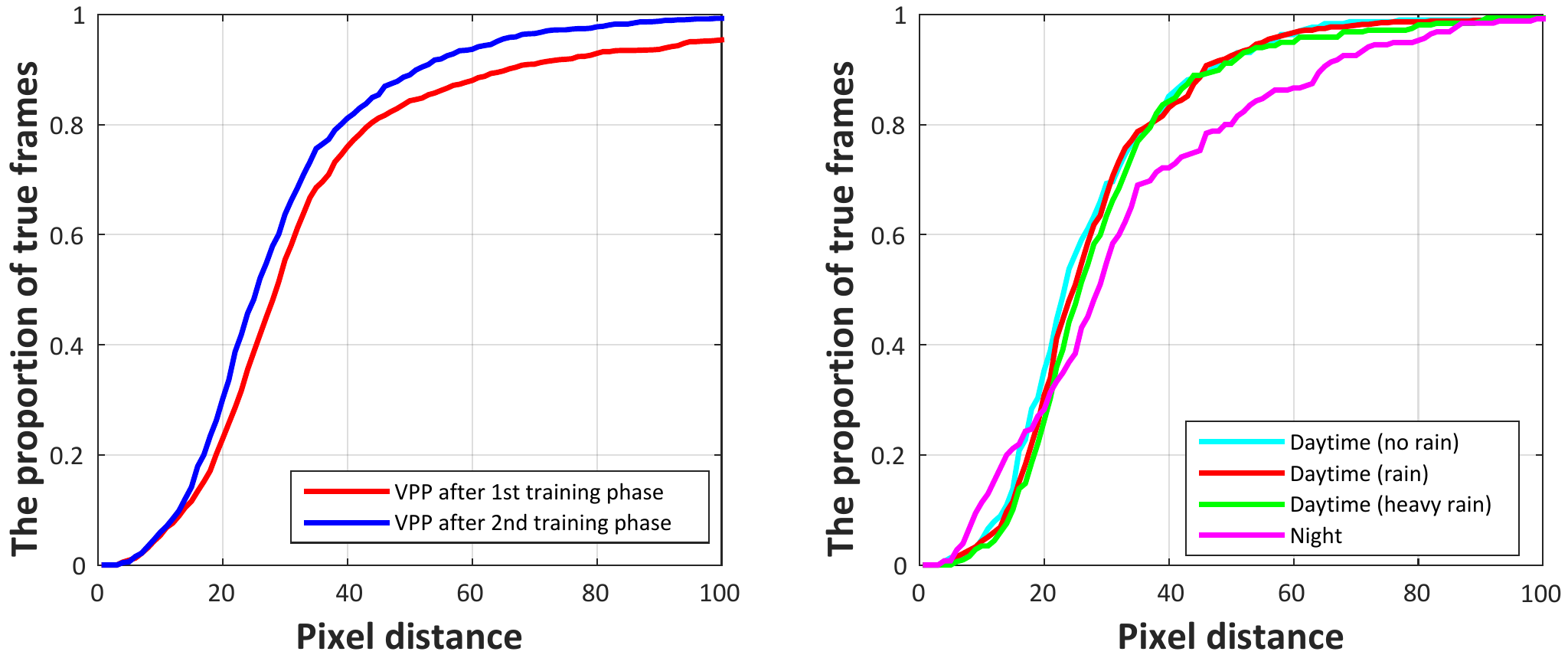}}\hspace{2mm}
	\end{center}
	\vspace{-5mm}
	\caption{Evaluation on the VPP task.}
	\label{fig:vp_eval}
	\vspace{-5mm}
\end{figure}

In the case of a VP, we compare the VPP-only and 4-Task networks. In this manner, we can observe how the VPP is influenced by the lane and road marking detection. Moreover, we compare the performances of each scenario. Figure~\ref{fig:vp_eval} shows the experimental results. The left graph shows a comparison between two outputs: a prediction after the first phase and a prediction after the second phase. The prediction after the second phase is highly improved meaning that the VPP task gets help from lane and road marking detection tasks. The right graph shows the results of the prediction after the second phase for each scenario.

\section{Conclusions}
\label{sec:6.}
\vspace{-1mm}

In this work, we introduced lane and road marking benchmark that covers four scenarios: daytime (no rain, rain, heavy rain) and night conditions. We have also proposed a multi-task network for simultaneous detection and classification of lane and road markings, guided by a VP. The evaluation shows that the VPGNet model is robust under different weather conditions and performs in real-time. Furthermore, we have concluded that the VPP task enhances both lane and road marking detection and classification by enhancing activation of lane and road markings and the boundary of the roadway.


\section*{Acknowledgement}
\vspace{-1mm}

This work was supported by DMC R\&D Center of Samsung Electronics Co.

{\small
	\bibliographystyle{ieee}
	\bibliography{egbib}
}

\end{document}